\documentclass[11pt]{article}

\usepackage[margin=1in]{geometry}
\usepackage{microtype}
\usepackage{graphicx}
\usepackage{float}
\usepackage{subcaption}
\usepackage{booktabs}
\usepackage{amsmath}
\usepackage{amssymb}
\usepackage{mathtools}
\usepackage{xcolor}
\usepackage{colortbl}
\usepackage{multirow}
\usepackage{natbib}
\usepackage{hyperref}
\usepackage[capitalize,noabbrev]{cleveref}

\newcommand{\DeltaEM}{\ensuremath{\Delta_{\text{L4--L1}}}}
\newcommand{\kap}{\ensuremath{\kappa}}

\title{Does Verbose Chain-of-Thought Really Help?\\In-Distribution Evidence that Content, Not Length, Matters}
\author{%
  Wenlong Wang\\ Fin AI Research\\ \texttt{wenlong.wang@intercom.io}
  \and
  Fergal Reid\\ Fin AI Research\\ \texttt{fergal.reid@intercom.io}
}
\date{}

\begin{document}

\maketitle

% =============================================================================
% ABSTRACT
% =============================================================================
\begin{abstract}
Chain-of-thought (CoT) prompting improves large language model reasoning, but the source of this improvement remains contested: does CoT help because the intermediate steps carry useful semantic content, or because conditioning on more tokens provides additional computational affordance before the model must commit to an answer?
We bring two complementary lines of evidence to bear.
First, \emph{in distribution}: we repeatedly sample each model on the same question and pair a shorter with a longer of its \emph{own} natural generations that follow the \emph{same reasoning plan}---so nothing is rewritten and both traces are genuinely in-distribution; across 25 models the extra tokens leave accuracy essentially unchanged for every independently-trained reasoner, and a blind analysis of the surplus tokens shows that what gain exists elsewhere tracks validation- and checking-content, not verbosity per se.
Second, as a controlled intervention, we ask whether two traces expressing the \emph{same} semantic content---the same facts, operations, and intermediate values, verified through directed acyclic graph (DAG) equivalence---produce different outcomes when one is more verbose, using a dual-validator design (an algorithmic DAG-informed validator paired with a more flexible LLM judge) across four targets and eight benchmarks with number-redacted completion and stratified bootstrap confidence intervals.
Verbose traces do improve accuracy---25 of 32 benchmark--target cells are positive under at least one validator---but the effects are modest (typically 1--4 percentage points) and depend on the \emph{quality} of the verbose prose, not merely its length.
Under maximum numerical redaction, where every number is replaced by a placeholder, the verbose-helps effect is \emph{amplified} (median $3.24\times$ across four arithmetic benchmarks), and length-matched non-reasoning filler recovers none of it.
Both lines converge: what matters is what the extra tokens \emph{do}---the reasoning and validation content they carry---not how many there are, a picture that neither a pure forward-pass-compute account nor a pure semantic-content account fully explains.
\end{abstract}

% =============================================================================
% 1. INTRODUCTION
% =============================================================================
\section{Introduction}
\label{sec:intro}

Chain-of-thought (CoT) prompting improves large language model reasoning by providing or eliciting intermediate reasoning steps before the final answer~\citep{wei2022chain, kojima2022large}.
Two accounts compete to explain \emph{why} these intermediate tokens help.
The \textbf{semantic-trace hypothesis} holds that CoT helps because it explicitly represents useful intermediate reasoning content---facts, operations, and intermediate values that the model conditions on when producing the answer.
The \textbf{forward-pass-computation hypothesis} holds that CoT helps primarily because additional tokens afford the model more serial computation before it must commit to an answer---whether those tokens are self-generated intermediate steps or externally supplied context~\citep{pfau2024dot, goyal2023pause}.
Disentangling these two accounts is increasingly urgent: recent work on looped transformers~\citep{giannou2023looped} and latent-space reasoning~\citep{hao2024coconut} deliberately trades away human-readable intermediate steps in favour of raw computation, making it essential to understand how much of the CoT benefit comes from the semantic content of the trace versus the additional computation it provides.

\citet{jin2024impact} provided influential evidence bearing on this question.
They showed that lengthening reasoning steps in few-shot CoT demonstrations improves performance across eight reasoning benchmarks, and that even incorrect rationales can be effective if they preserve sufficient inference length---a finding consistent with the forward-pass account.
However, their design modifies the few-shot \emph{demonstrations} rather than the target problem's trace, which also shifts the model's generation distribution; their expansion strategies (``Read the question again'', ``Self-Verification'') are researcher-designed elaborations that may contribute useful structure beyond mere token count; and their evaluation uses a single validator per benchmark without equivalence verification on the rewritten traces themselves.
Our experiments, which control for these factors, point to a more mixed picture.

Despite this body of work, several challenges remain in the effort to disentangle the semantic-trace and forward-pass-computation accounts.
First, any trace-length intervention must verify that the rewritten traces preserve the same semantic reasoning structure as the original---the same facts, operations, intermediate values, and dependencies---rather than subtly adding or removing reasoning content along with verbosity.
Second, any rewrite inevitably shifts the trace out of the distribution the target model would produce under standard CoT prompting (e.g., ``Let's think step by step''); in a diagnostic pilot using lossless speculative decoding, we found 100\% first-token rejection on all 335 verbose rewrites---even when the drafter and target shared the same model weights but differed only in conditioning prompt---confirming that supplied traces are out-of-distribution regardless of their semantic content.
Third, evaluation itself introduces bias: substring-match validators, commonly used in CoT studies, can exhibit false-positive rates of 50--64\% when audited against LLM judges (as we found in an early phase of this project), and non-arithmetic benchmarks have had no validator-uniform large-sample evaluation because algorithmic validators are structurally undefined for prose-based tasks.

We address these gaps with four contributions, organised as two complementary legs---an in-distribution test and a controlled-intervention battery---that converge on the same mechanism.

\begin{enumerate}
\item \textbf{In-distribution (same-plan).} To avoid supplying any externally-written trace, we sample each model repeatedly on the same question (temperature sampling, $T = 0.7$) and, from its \emph{own} natural generations, select a shorter (concise) and a longer (verbose) one that commit to the \emph{same reasoning plan}---so both are genuinely in-distribution and differ only in how many tokens they spend on a fixed approach. Measuring $\Delta$, the verbose-minus-concise accuracy difference, across 25 models, length helps in only two places---DeepSeek-R1-Distill's specific released weights (on both MATH-500 and GSM8K) and non-reasoners on GSM8K---while for every other independently-trained reasoner $\Delta \approx 0$; a blind extra-token analysis attributes what gain exists to validation/checking content rather than token count. Because nothing is rewritten, this leg \emph{closes the out-of-distribution objection} to the intervention leg.

\item \textbf{Methodological.} We introduce a \emph{dual-validator design} pairing an algorithmic DAG-informed equivalence validator (E2, defined on arithmetic benchmarks) with a local LLM judge (E3, defined on all benchmarks), calibrated at $\kap = 0.586$ against a frontier judge.
Number-redacted completion blocks answer-copying and enables the redaction-amplification analysis below.

\item \textbf{Empirical.} A partial replication of the verbose-helps finding across 8 benchmarks $\times$ 4 targets $\times$ 2 validators: 25 of 32 cells are positive under at least one validator, with typical effect sizes of 1--4\,pp under the LLM judge---modest, and many individual cells have bootstrap CIs crossing zero.
The effect is highly sensitive to the validation procedure (E2 estimates are $3{-}4\times$ larger than E3), to the rewrite source (shared-rewriter effects exceed self-rewrite effects in 14 of 18 sign-discrepant cells), and to the redaction regime.

\item \textbf{Mechanism-constraining.} Under maximum numerical redaction, the verbose-helps effect is \emph{amplified} ($3.24\times$ median across four arithmetic benchmarks) rather than eliminated---the diagnostic signature of a prose-quality mechanism rather than a numerical-scaffolding or pure-token-count effect.
Two further patterns constrain the mechanism: \emph{asymmetric validator disagreement} (E2 and E3 select on opposite sides of the verbosity axis) and \emph{rewriter-quality dependence} (different prose authors at comparable verbosity yield different effect sizes).
\end{enumerate}

\Cref{sec:related} reviews related work.
\Cref{sec:methods} describes the experimental pipeline.
\Cref{sec:experiments} presents results.
\Cref{sec:discussion} interprets the findings, and \Cref{sec:limitations} discusses limitations.

% =============================================================================
% 2. RELATED WORK
% =============================================================================
\section{Related Work}
\label{sec:related}

\paragraph{CoT and reasoning length.}
Scratchpad-style intermediate computation improves performance on algorithmic tasks~\citep{nye2021scratchpads}, and few-shot CoT prompting yields large gains on arithmetic, commonsense, and symbolic reasoning~\citep{wei2022chain, kojima2022large}.
Complexity-based prompting selects demonstrations with more reasoning steps and reports improvements on multi-step tasks~\citep{fu2022complexity}.
Most directly related, \citet{jin2024impact} manipulate reasoning-step length in CoT demonstrations, reporting that lengthening steps improves performance and that even incorrect rationales can remain effective if they preserve sufficient length.
However, prompt-based length interventions change the distribution over generated traces, not just the number of tokens, making it difficult to isolate whether verbosity itself causally improves the final answer.

\paragraph{Extra tokens as computation.}
\citet{pfau2024dot} show that meaningless filler tokens can provide computational benefit on algorithmic tasks, though learning to use them requires dense supervision.
\citet{goyal2023pause} find that learnable pause tokens improve performance when models are pretrained with such delays.
Budget-forcing methods lengthen reasoning by appending continuation tokens~\citep{muennighoff2025s1}, while \citet{ghosal2025thinking} find non-monotonic effects---additional thinking initially helps but eventually degrades performance.
These studies motivate the forward-pass hypothesis but typically require training or use synthetic tasks.

\paragraph{Faithfulness and counterfactual analysis.}
CoT traces can systematically rationalise biased predictions~\citep{turpin2023language}.
\citet{lanham2023measuring} evaluate faithfulness by truncating, corrupting, and paraphrasing CoT traces.
\citet{madaan2023makes} find that surface patterns in CoT demonstrations can matter more than exact symbols.
\citet{garcia2026lastword} identify answer-placement confounds in corruption-based faithfulness tests.
These motivate our dual-validator design and use of number-redacted completion to prevent answer copying.

% =============================================================================
% 3. METHODS
% =============================================================================
\section{Methods}
\label{sec:methods}

\subsection{Overview: Two Legs}
\label{sec:overview}

Both legs estimate the same quantity, $\Delta = \mathrm{acc(verbose)} - \mathrm{acc(concise)}$: a paired contrast between a more verbose and a more concise reasoning trace that express the \emph{same} underlying reasoning, so that any accuracy difference is attributable to verbosity rather than content.
They differ in how each concise/verbose pair is obtained and how same-content is verified.
\emph{Leg~1} (in-distribution; \Cref{sec:samedag}) is our primary design: it pairs a model's \emph{own} natural samples that commit to the same reasoning plan, so neither trace is rewritten.
\emph{Leg~2} (controlled intervention; \Cref{sec:rewrite}) rewrites a model's trace into matched concise and verbose forms while holding a parsed computation graph fixed, trading in-distribution fidelity for tighter control over content and verbosity.
The two legs share one equivalence instrument, the LLM judge \textbf{E3} (Qwen3-Next-80B-A3B-Instruct-FP8, served via vLLM with xgrammar-constrained JSON output), which reads two number-redacted traces and returns a verdict $\in \{\textit{equivalent}, \textit{not\_equivalent}, \textit{ambiguous}\}$ with a confidence score; E3 is defined on all eight benchmarks.
Number redaction (masking numeric tokens before judging or completion) and anchor/question-clustered bootstrap confidence intervals are used throughout.
Leg~2 adds a second, \emph{algorithmic} validator (\textbf{E2}, arithmetic-only); Leg~1 uses no algorithmic validator, relying on E3 plus an independent frontier-judge cross-check.

\subsection{Leg 1: In-Distribution Same-Plan Pairing}
\label{sec:samedag}

This leg works entirely with a model's \emph{own} natural samples, so both traces are in-distribution---directly answering the objection that supplied rewrites (Leg~2) are out-of-distribution.
For each (model, question) we draw many natural rollouts ($T = 0.7$, top-$p = 0.95$); because sampled traces vary in length, this yields, for most questions, both short and long generations of the same problem from the model's own distribution.
From these we form pairs of a \emph{concise} and a \emph{verbose} generation (generated-token ratio $\geq 1.5$) that commit to the \emph{same reasoning plan}; nothing is rewritten or externally supplied.
We estimate $\Delta$ with a question-clustered bootstrap ($B = 10{,}000$).
Fixing the plan removes the content confound that makes naively-sampled length look harmful: longer traces are disproportionately the ones a model resorts to on harder attempts, so the raw length--accuracy correlation is negative (``overthinking'').

\paragraph{Verifying same-plan with two judges.}
This leg uses \emph{no} algorithmic (E2) validator; same-plan is established by LLM judges on number-redacted text, in two stages chosen to be both cheap and robust.
First, the local judge E3 confirms every nominated pair; on GSM8K this is preceded by a value-agnostic structural pre-filter---candidate pairs must share a computation-graph signature (a Weisfeiler--Lehman hash over the arithmetic op-graph, matching traces that perform the same operations in the same dependency structure regardless of the literal numbers).
On MATH-500---symbolic, with no arithmetic graph to fingerprint---this pre-filter is inert, so \emph{E3 alone} decides same-plan, and it is therefore the entire instrument on the benchmark that carries the result.
Second, to show the finding does not rest on that one local judge, an independent frontier judge (Claude Opus 4.8) re-runs E3's exact blind task across all 25 MATH-500 cells.
For cost efficiency it judges only the pairs that can move the estimate: all mixed-correctness pairs---where the concise and verbose samples disagree on correctness, and which alone shift $\Delta$---plus a per-cell same-correctness sample to anchor the denominator (6{,}065 pairs total).
Opus is not treated as ground truth; the comparison tests whether $\Delta$ survives the judge swap (\Cref{sec:samedag-results}).

We evaluate 25 models spanning DeepSeek-R1-Distill variants, independently-trained reasoners (Qwen3, Phi-4-reasoning-plus, Magistral, gpt-oss, Sky-T1, Bespoke-Stratos, OpenR1, OpenThinker), and non-reasoning models.
MATH-500 is the spine (all 25 models, $n \approx 300\text{--}440$ per cell) and is scored with a LaTeX-normalised grader; GSM8K is a second benchmark (${\sim}13$ models).

\paragraph{What the extra tokens do.}
To ask whether any gain comes from token count or from content, a blind labeller (outcome-masked) tags the surplus of each verbose trace over its concise partner with a 10-label taxonomy distinguishing pure elaboration (\texttt{elaborate/restate}) from validation reasoning (\texttt{verify/check}, \texttt{plug-in/test}).
We report within-judge win-versus-loss label-presence contrasts---robust to a labeller's absolute biases because the contrast is internal to each judge---cross-checked by a second labeller (Opus) and by two model-free lexical analyses.

\subsection{Leg 2: Controlled-Intervention Rewrites}
\label{sec:rewrite}

Leg~2 trades in-distribution fidelity for control: it rewrites a target's reasoning trace into matched concise and verbose forms while holding the underlying computation fixed, enabling manipulations the in-distribution leg cannot perform---prose-quality and rewriter-source swaps, maximum numerical redaction, and coverage of non-arithmetic benchmarks.

\paragraph{Reasoning-trace DAGs.}
\label{sec:dags}
For arithmetic traces we extract a directed acyclic graph (DAG) in which each node is an intermediate computation (e.g., ``$12 + 8 = 20$'') and edges encode step-to-step dependencies.
Two traces are \emph{DAG-equivalent} if they contain the same computational nodes in the same dependency order with identical intermediate values---regardless of how concisely or verbosely each step is expressed.
This is a stronger, value-aware criterion than the value-agnostic signature used in Leg~1: by verifying that a concise and a verbose rewrite share the same DAG before measuring accuracy, we isolate verbosity from reasoning content.
DAG equivalence captures computational content but not expository structure (e.g., discourse markers or subgoal framing); E3 provides a complementary check on those broader features.

\begin{figure}[t]
\centering
\includegraphics[width=\textwidth]{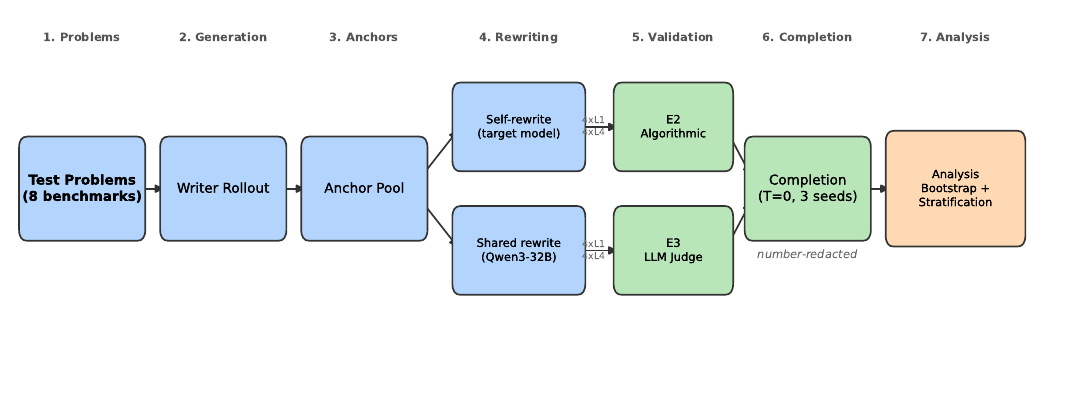}
\caption{Leg~2 (controlled-intervention) pipeline.
Each target model generates a reasoning trace for every test problem; parseable traces form the anchor pool.
Each anchor is rewritten at two verbosity levels (L1 concise, L4 verbose) by two sources (self-paraphrase and shared rewriter), validated by both an algorithmic (E2) and LLM-judge (E3) validator, then used to prompt number-redacted target completions.
Results are analysed via paired bootstrap with $2 \times 2 \times 2$ stratification.}
\label{fig:pipeline}
\end{figure}

The pipeline (\Cref{fig:pipeline}) implements six design pillars that address limitations of prior verbose-CoT evaluations:

\begin{enumerate}
\item \textbf{Anchor generation.} For each benchmark, every test-set question is presented to the target model with a chain-of-thought prompt. The model's reasoning trace is parsed into a DAG; every parseable trace becomes an anchor regardless of final-answer correctness, enabling stratified analysis by anchor correctness.

\item \textbf{Multi-sample rewrites.} For each anchor, we generate 4~L1 (concise) and 4~L4 (verbose) candidate rewrites per source, yielding ${\sim}98\%$ pairing success after per-rewrite validation.

\item \textbf{Two rewrite sources.} \emph{Self-rewrites} are produced by the evaluation target itself; \emph{shared rewrites} are produced by a canonical strong rewriter (Qwen3-32B).

\item \textbf{Dual validation.} Both E2 (algorithmic) and E3 (LLM judge) are applied to every rewrite.

\item \textbf{Number-redacted completion.} The target model sees the question and rewrite with numeric tokens replaced by \texttt{<REDACTED>}, followed by ``The answer is'', generating at $T{=}0$ across 3 seeds.
This forces the target to compute the answer rather than echo pre-computed values.

\item \textbf{Stratified analysis.} We report $\DeltaEM$ exact-match accuracy across a $2 \times 2 \times 2$ stratification (source $\times$ anchor-correctness $\times$ validator) with 1{,}000-sample paired bootstrap CIs, resampling at the anchor level to respect the dependence structure across multiple rewrites per anchor.
\end{enumerate}

\noindent \Cref{app:example} provides a concrete worked example showing the full pipeline for one GSM8K problem, including the anchor trace, L1/L4 rewrites, number-redacted completion prompt, and judge rubric.

\paragraph{E2: algorithmic validator.}
A 6-feature logistic regression informed by parsed DAG structure, calibrated at threshold $\tau^* = 0.48$ on 500 calibration + 160 hold-out LLM-as-judge labels (Claude Sonnet).
Features include substring-match pass, topological-order validity, answer-leakage check, token-ratio, prose--equation alignment, and topological-signature match.
The inclusion of token-ratio as a feature means E2 is not a pure graph-isomorphism check; it is a learned equivalence classifier whose verbosity sensitivity is analysed in \Cref{sec:asymmetry}.
E2 is defined only on arithmetic benchmarks (Tiers~A and B); it is structurally undefined for non-arithmetic tasks.

\subsection{Calibration}

On 5{,}562 GSM8K rows jointly labelled by both Sonnet and the local Qwen judge, binary Cohen's $\kap = 0.586$ with 88.3\% raw agreement (\Cref{fig:calibration} in Appendix).
Equivalent-class recall is 95.8\%.
Adding a \texttt{REDACTED}-aware clarification block to the rubric raised $\kap$ from 0.44 to 0.59.
We interpret this as moderate-to-substantial agreement~\citep{landis1977measurement}, sufficient for cross-validator comparison but below the level where single-judge results should be trusted as ground truth.

\subsection{Benchmarks and Targets}

The in-distribution leg's 25 models are listed in \Cref{sec:samedag}.
Leg~2 evaluates four target models, selected to vary across scale, architecture, and model family: Qwen3-4B (4B, dense), Qwen3-32B (32B, dense), Qwen3-30B-A3B (mixture-of-experts, 3B active parameters), and OLMo-2-7B (AllenAI, 7B, dense) as a cross-family control; three of four belong to the Qwen family, a deliberate trade-off discussed in \Cref{sec:limitations}.
It spans eight benchmarks grouped into three tiers by algorithmic-validator applicability (\Cref{tab:tiers}).

\begin{table}[t]
\centering
\small
\caption{Leg~2 benchmark tiers and E2 validator applicability.}
\label{tab:tiers}
\begin{tabular}{@{}lp{3.8cm}c@{}}
\toprule
Tier & Benchmarks & E2 \\
\midrule
\rowcolor{gray!15} A (arithmetic) & GSM8K, MATH-500,\newline MultiArith, SVAMP & \checkmark \\
B (algebraic MCQ) & AQuA & partial \\
\rowcolor{gray!15} C (non-arithmetic) & Letter, Coin, StrategyQA & --- \\
\bottomrule
\end{tabular}
\end{table}

% =============================================================================
% 4. EXPERIMENTS
% =============================================================================
\section{Experiments}
\label{sec:experiments}

\subsection{Same-Plan Length Effect (In-Distribution)}
\label{sec:samedag-results}

We begin with the in-distribution leg, which is immune to the out-of-distribution objection that the rewrite-based results below must address.
\Cref{tab:samedag} reports the result.
Holding the reasoning plan fixed removes a confound that is easily mistaken for ``overthinking.''
In naively-sampled data, a model's longer traces are disproportionately its harder---and more error-prone---attempts, producing a spurious negative length--accuracy correlation (qwen3-4B think: $-0.252$).
Once the plan is held constant, so that paired traces differ only in rendering length, this signal collapses to ${\approx}0$.
For every independently-trained reasoner, the amount of reasoning carries no accuracy signal once the plan is fixed (pooled MATH-500 $\Delta = -0.005$ [$-0.010$, $-0.000$]).

Length genuinely helps in only two places, and the split is by \emph{post-training}, not family, scale, or teacher.
DeepSeek-R1-Distill's specific released weights help on both MATH-500 ($+0.019$ [$+0.010$, $+0.027$]) and GSM8K (deepseek-8B $+0.046$, $n = 1{,}033$); independent clones on the same base, teacher, and recipe are $\Delta \approx 0$, so this is a weights idiosyncrasy.
Non-reasoning models help on GSM8K (mistral-7B $+0.079$; qwen3-0.6B-nothink $+0.046$) but not on MATH-500 ($+0.004$ [$-0.016$, $+0.023$]).

\begin{table}[t]
\centering
\small
\caption{Same-plan $\Delta = \mathrm{acc(verbose)} - \mathrm{acc(concise)}$ on MATH-500 (family-pooled, LaTeX-normalised grader), with an independent Opus same-plan judge for comparison. CIs are question-clustered bootstrap ($B = 10{,}000$).}
\label{tab:samedag}
\resizebox{\columnwidth}{!}{%
\begin{tabular}{@{}lcc@{}}
\toprule
Family (arms) & $\Delta$ (E3 judge) & $\Delta$ (Opus judge) \\
\midrule
DeepSeek-R1-Distill (5) & $+$.019 [$+$.010, $+$.027] & $+$.032 [$+$.022, $+$.043] \\
Independent reasoners (13) & $-$.005 [$-$.010, $-$.000] & $-$.007 [$-$.012, $-$.001] \\
Non-reasoners (5)$^\dagger$ & $+$.004 [$-$.016, $+$.023] & $-$.024 [$-$.046, $-$.002] \\
\bottomrule
\end{tabular}%
}
\vspace{2pt}\\
{\footnotesize $^\dagger$mistral-7B's underpowered MATH-500 cell is excluded; its well-powered GSM8K cell ($+0.079$) is retained. The Opus cross-check pools span 5\,/\,14\,/\,5 arms.}
\end{table}

\paragraph{What the extra tokens do.}
A blind extra-token analysis resolves the mechanism.
Pure verbosity (\texttt{elaborate/restate}) \emph{never} discriminates a winning from a losing pair under either labeller, in any family---the single most judge-robust finding.
Validation reasoning does: for DeepSeek both \texttt{plug-in/test} and \texttt{verify/check} are significantly enriched in wins under both labellers, and \texttt{verify/check} replicates in independent reasoners.
Two model-free analyses (a curated reasoning-cue lexicon and Fightin'-Words log-odds on the surplus tokens) reproduce the directional signal with no LLM in the loop.
Where length helps, it is what the tokens do, not how many there are.

\paragraph{Judge robustness.}
On MATH-500 the E3 judge is the entire same-plan instrument, so we re-ran its task with an independent frontier judge (Claude Opus 4.8) across all 25 cells (6{,}065 pairs).
The split is judge-independent and the swap only sharpens it: the DeepSeek pool rises to $+0.032$ [$+0.022$, $+0.043$], independent reasoners stay ${\approx}0$, and the local judge proves the \emph{stricter} of the two (it rejected only $6.4\%$ of E3's DeepSeek confirmations while also confirming pairs E3 had rejected), ruling out a lax-judge inflation of the effect.

\subsection{Headline Replication}
\label{sec:headline}

\begin{figure}[t]
\centering
\includegraphics[width=\textwidth]{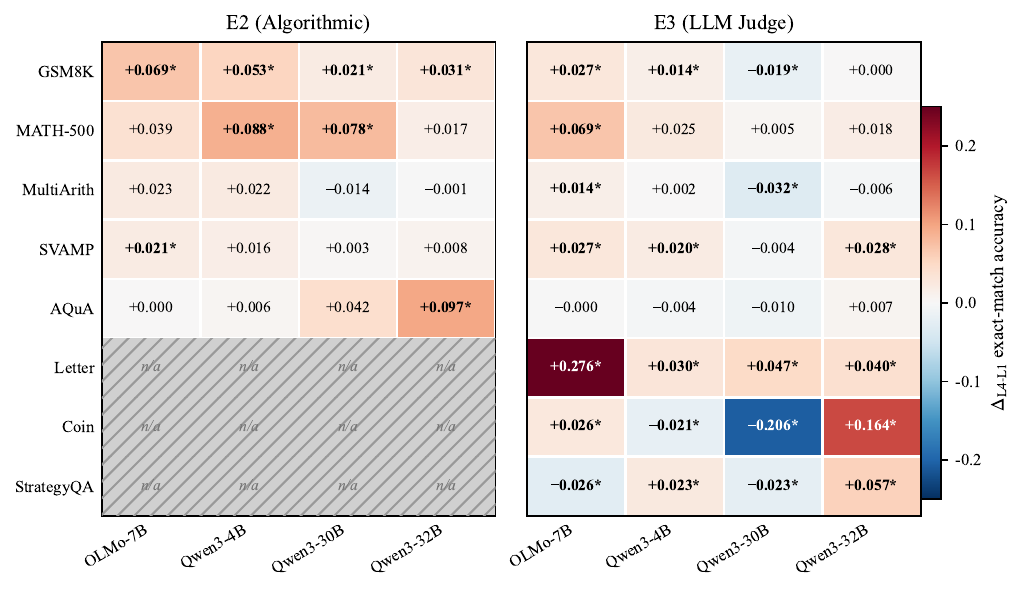}
\caption{$\DeltaEM$ (verbose L4 minus concise L1) across 8 benchmarks and 4 targets.
Left: E2 algorithmic validator (Tier~C undefined, shown as hatched).
Right: E3 LLM judge (all benchmarks).
Asterisks indicate 95\% bootstrap CIs excluding zero.
Colour scale is symmetric around zero; red indicates verbose-helps and blue indicates verbose-hurts.}
\label{fig:heatmap}
\end{figure}

\Cref{fig:heatmap} presents the main result: $\DeltaEM$ across 8 benchmarks $\times$ 4 targets under both validators.
The E2 panel (left) shows uniformly positive effects on Tier~A arithmetic benchmarks, with GSM8K significant across all four targets ($+0.021$ to $+0.069$) and AQuA $\times$ Qwen3-32B reaching $+0.097$, though the latter rests on only $n = 8$ paired samples and should be treated as indicative.
Tier~C benchmarks (Letter, Coin, StrategyQA) are undefined under E2.

The E3 panel (right) reveals a more nuanced picture.
Tier~A effects are smaller: GSM8K ranges from $-0.019$ (Qwen3-30B) to $+0.027$ (OLMo), and typical Tier~A magnitudes fall in the 1--4\,pp range.
However, E3 enables evaluation on Tier~C, where Letter shows the largest effects in the study ($+0.276$ for OLMo, $+0.030$ to $+0.047$ for Qwen targets), while Coin and StrategyQA are mixed.

\Cref{tab:sign} summarises the sign-agreement pattern across all 32 cells, classified by point-estimate sign (treating $\pm 0.000$ as null).
Under the criterion of \emph{both} validators showing a positive point estimate, 12 of 20 Tier~A/B cells replicate in direction, though several have bootstrap CIs crossing zero (\Cref{tab:e2full,tab:e3full}).
Under the broader criterion of \emph{at least one} validator positive, 25 of 32 cells agree with Jin et~al.'s reported direction.
The 5-cell gap (E2-positive but E3-null or negative) is analysed in \Cref{sec:asymmetry}.

\begin{table}[t]
\centering
\small
\caption{Sign-agreement breakdown across 32 (benchmark $\times$ target) cells, classified by point-estimate sign.}
\label{tab:sign}
\begin{tabular}{@{}lrl@{}}
\toprule
Pattern & Count & Interpretation \\
\midrule
E2+ and E3+ & 12 & Both validators positive \\
E2+ only (E3 ${\leq}$\,0) & 5 & Validator disagreement \\
E2 undef., E3+ & 8 & Tier~C (E3 only) \\
Both ${\leq}$\,0 or null & 7 & Null or negative \\
\midrule
Total & 32 & \\
\bottomrule
\end{tabular}
\end{table}

\subsection{Asymmetric Validator Disagreement}
\label{sec:asymmetry}

\begin{figure}[H]
\centering
\includegraphics[width=\columnwidth]{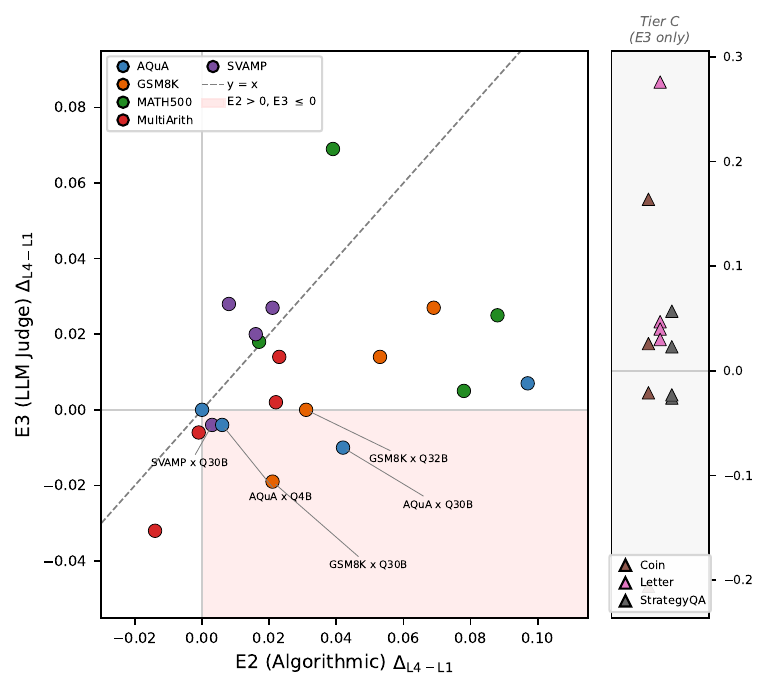}
\caption{E2 versus E3 $\DeltaEM$ per cell.
Points in the shaded region (E2 $> 0$, E3 $\leq 0$) indicate cells where the two validators disagree on the sign of the verbose-helps effect.
Five such cells are annotated.
Tier~C points (right strip) have no E2 value.
Most points fall below the $y = x$ diagonal, indicating the two validators produce systematically different magnitude estimates.}
\label{fig:scatter}
\end{figure}

\Cref{fig:scatter} plots E2 versus E3 $\DeltaEM$ for all cells where both validators are defined.
An asymmetry emerges: 5 cells show E2-positive but E3-null or negative effects, with no clear cases of the reverse pattern among Tier~A/B cells.
The affected cells are concentrated on Qwen3-30B and Qwen3-32B:

\begin{itemize}
\item GSM8K $\times$ Qwen3-30B: E2 $= +0.021$, E3 $= -0.019$
\item GSM8K $\times$ Qwen3-32B: E2 $= +0.031$, E3 $= +0.000$
\item AQuA $\times$ Qwen3-30B: E2 $= +0.042$, E3 $= -0.010$
\item AQuA $\times$ Qwen3-4B: E2 $= +0.006$, E3 $= -0.004$
\item SVAMP $\times$ Qwen3-30B: E2 $= +0.003$, E3 $= -0.004$
\end{itemize}

The two validators select on \emph{opposite} sides of the verbosity axis.
E2 systematically over-rejects verbose L4 rewrites: on Tier~A benchmarks, E2 acceptance is 18--32\,pp lower for L4 than L1 (e.g.\ GSM8K shared: L1 28.8\%, L4 12.9\%).
Conversely, E3 over-accepts L4: the LLM judge finds verbose prose easier to certify as equivalent, with L4 acceptance 2--34\,pp higher than L1.
Neither validator is unbiased toward verbosity: E2 favours concise rewrites, E3 favours verbose ones, and their disagreement is structural rather than random.

\paragraph{Intersection analysis.}
To disentangle selection from measurement, we recompute $\DeltaEM$ on the subset of rewrites accepted by \emph{both} validators (E2$\cap$E3).
Across 43 Tier~A/B cells with defined intersection pools, 19 preserve the sign of the unfiltered $\DeltaEM$ and 5 remain CI-significant in the same direction, with median magnitude attenuation of ${\sim}50\%$.
Crucially, none of the 14 apparent sign-changes represents a real reversal: all have intersection point estimates within $\pm 0.02$ of zero.
The intersection result confirms that the strongest verbose-helps cells (e.g.\ GSM8K $\times$ Qwen3-4B self: $+0.013$ [$+0.005$, $+0.023$]) survive dual-validator filtering, while marginal cells collapse to noise rather than flipping sign.

\subsection{Rewriter-Quality Dependence}
\label{sec:rewriter}

\begin{figure}[H]
\centering
\includegraphics[width=0.7\columnwidth]{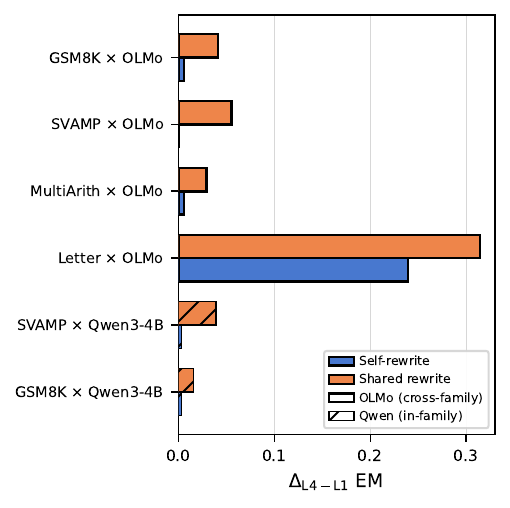}
\caption{Self-rewrite versus shared-rewrite $\DeltaEM$ (E3 validator) for selected cells.
OLMo (cross-family) shows near-zero self-rewrite effects but substantial shared-rewrite effects.
Note that shared-L4 rewrites are 17--45\% longer at the median; see \Cref{sec:limitations} for discussion of this confound.}
\label{fig:rewriter}
\end{figure}

We decompose $\DeltaEM$ by rewrite source (self-paraphrase versus shared Qwen3-32B rewriter).
Across 52 (benchmark $\times$ target $\times$ validator) cells where both sources are defined, 18 (35\%) show a sign disagreement between self and shared rewrites.
Of these 18, \textbf{14 have shared $>$ self in magnitude}, with only 4 in the reverse direction.
The effect is most pronounced on the cross-family target OLMo-2-7B (\Cref{fig:rewriter}):

\begin{itemize}
\item GSM8K: self $\Delta = +0.006$, shared $\Delta = +0.041$ ($7\times$)
\item SVAMP: self $\Delta = +0.000$, shared $\Delta = +0.055$
\item MultiArith: self $\Delta = +0.006$, shared $\Delta = +0.029$ ($5\times$)
\item Letter: self $\Delta = +0.239$, shared $\Delta = +0.314$ ($1.3\times$)
\end{itemize}

\paragraph{Length confound.}
Shared-L4 rewrites are systematically longer than self-L4 (median 17--45\% longer across benchmarks), so the shared advantage cannot be attributed to prose quality alone---it is confounded with additional tokens.
However, the within-source L4$-$L1 token gap is not always larger for the shared source.
For example, Qwen3-30B's self L4$-$L1 gap on GSM8K is 486 tokens versus the shared gap of 363 tokens, yet the shared rewriter still produces a larger $\DeltaEM$.
The cleanest resolution of this confound comes from the redaction-amplification analysis below (\Cref{sec:redaction}), which holds token count constant while stripping numerical content.

\subsection{Redaction Amplification}
\label{sec:redaction}

If verbose-helps were primarily an answer-extraction effect---verbose traces exposing more numerical values for the target to copy---then the effect should weaken when numerical information is unavailable.
We test this with two analyses, each using a distinct redaction regime from the main pipeline (which replaces all numeric tokens in the rewrite with \texttt{<REDACTED>}).

\paragraph{Answer-leak stratification (unredacted completions).}
In a separate analysis using completions where the target model sees the full unredacted rewrite (all numbers visible), we split each cell by whether the rewrite text contains the gold answer as a standalone numeric token (60--80\% of rewrites do, because the rewriter often re-derives the gold value as an intermediate computation).
In 29 of 32 Tier~A/B cells, $\DeltaEM$ is \emph{larger} on the no-leak subset than on the leak subset---the opposite of what the answer-extraction account predicts.
The mechanism is ceiling saturation: when the gold answer is directly visible in both L1 and L4, both conditions approach near-perfect accuracy and within-anchor $\Delta \to 0$; when it is absent, the target must reason from intermediate prose, and verbose L4 provides more useful structure to reason from.

\paragraph{Maximum numerical redaction.}
In a more aggressive regime than the main pipeline, we replace \emph{every} numeric token in each rewrite with \texttt{<NUM>} and recomplete (200-anchor subsample, $T{=}1.0$, seed=0).
Across four numeric benchmarks (GSM8K, MATH-500, MultiArith, SVAMP), 30 of 32 cells preserve sign and 28 of 32 remain CI-significant, with a median amplification of $3.24\times$ relative to the original $\DeltaEM$.
The amplification is substantial: for instance, SVAMP $\times$ Qwen3-4B (self) rises from $\Delta = +0.005$ (within noise) to $+0.226$ [$+0.174$, $+0.288$]; OLMo on MultiArith (shared) rises from $+0.169$ to $+0.547$.
AQuA is excluded from this analysis because its multiple-choice format degenerates when answer-option labels are redacted.

Together, these two analyses suggest that the verbose-helps effect arises from prose-level reasoning structure rather than numerical scaffolding.
Under maximum redaction, terse L1 rewrites collapse to near-uninformative placeholder sequences while verbose L4 retains entity relationships, action structure, and reasoning narrative that the target can condition on.
However, we note that maximum redaction creates an aggressive OOD shift, so the amplified $\DeltaEM$ should be treated as a stress-test signal rather than a calibrated effect size.

\paragraph{Converging evidence from a filler condition.}
A separate oracle-trace experiment (\Cref{app:pilot}) provides independent evidence from the opposite direction.
Using canonical reasoning graphs extracted from gold solutions and rewritten into four verbosity levels by Qwen3-32B, we included a \emph{filler} condition: the concise L1 trace padded with fluent but non-reasoning text to match the L4 token count.
Across 6 models from 4 families, filler accuracy (${\sim}119$\,words) clusters near L1 rather than L3 or L4 (filler$-$L1 ranges from $-0.025$ to $+0.049$), while L3$-$filler at comparable length (${\sim}139$\,words) is significantly positive on every model ($+2$ to $+25$\,pp).
The same experiment also reveals that verbose-helps is domain-dependent: on StrategyQA's commonsense reasoning, 2 of 4 model families show verbose CoT \emph{hurts} accuracy, unlike the monotone benefit observed on mathematical benchmarks.

\subsection{Tier~C: Non-Arithmetic Benchmarks}
\label{sec:tierc}

E3 enables evaluation on Tier~C benchmarks, where E2 is structurally undefined.
Letter-concatenation shows the strongest verbose-helps signal: all four targets are significantly positive, with OLMo reaching $+0.276$ ($n = 457$).
Coin-flip is mixed (2 of 4 positive), with a notable outlier: Qwen3-30B shows $\DeltaEM = -0.206$ ($n = 487$), one of the largest negative effects in the study.
StrategyQA is also mixed (2 of 4 positive: Qwen3-32B $+0.057$, Qwen3-4B $+0.023$; OLMo and Qwen3-30B slightly negative).

\paragraph{Cross-family judge replication.}
To test whether Tier~C findings depend on the Qwen-family judge, we re-run all three Tier~C benchmarks under a non-Qwen judge (Yi-1.5-9B-Chat, 01.AI).
Individual-rewrite agreement between the two judges is low (mean $\kap = 0.29$), driven by Yi being substantially more lenient (90--95\% acceptance versus Qwen's 60--76\%).
However, the within-anchor $\DeltaEM$ contrast is robust to judge-swap: all 24 (benchmark $\times$ target $\times$ source) cells are sign-matched between the two judges, and 18 of 24 differ by less than 0.01 in absolute magnitude.
The key results---Letter $\times$ OLMo at $+0.294$ (Yi) versus $+0.314$ (Qwen), and Coin $\times$ Qwen3-30B at $-0.199$ (Yi) versus $-0.236$ (Qwen)---replicate under the cross-family judge.
Yi's extra-accepted rewrites distribute approximately uniformly across L1 and L4 within each anchor, so they do not shift the within-anchor contrast.

% =============================================================================
% 5. DISCUSSION
% =============================================================================
\section{Discussion}
\label{sec:discussion}

\paragraph{Constraining the mechanism.}
Our two legs are complementary by construction: the \emph{intervention} leg supplies rewritten traces as input context (testing whether \emph{conditioning on} verbose reasoning helps), while the \emph{in-distribution} leg (\Cref{sec:samedag-results}) uses only the model's \emph{own} natural samples at a fixed plan, so neither's conclusion rests on supplied prose.
Both reach the same result---raw length, holding the plan constant, does essentially nothing for independently-trained reasoners, and what gain exists tracks validation content---so the two legs are not vulnerable to either's single weakness.
A \emph{purely length-based account}---that the benefit comes solely from more tokens---is harder to reconcile with three independent findings: the redaction-amplification result (\Cref{sec:redaction}), where the verbose-helps effect \emph{grows} rather than shrinking under maximum numerical redaction; an oracle-trace experiment (\Cref{app:pilot}) in which length-matched non-semantic filler tokens failed to recover the CoT benefit; and, in-distribution, the same-plan extra-token analysis, in which pure verbosity never discriminates a winning pair from a losing one.
The observed $3.24\times$ amplification suggests that the mechanism depends on the prose-level reasoning structure that L4 retains and L1 loses under redaction.
A \emph{pure semantic-content account} is also partially disfavoured, because a real 1--4\,pp verbose-helps effect persists under the LLM judge when the verbose trace is written by a competent rewriter---semantic equivalence (DAG-verified) is held constant, yet verbosity still helps.
The best-supported account is \emph{content-driven with prose-quality as a multiplier}: CoT helps primarily through semantic content, but competent verbose elaboration adds a further prose-level affordance---entity relationships, discourse structure, redundant cues---that the target model can condition on.
We note that this does not rule out richer forward-pass accounts in which different token sequences provide different computational affordances; the boundary between ``semantically useful prose'' and ``computationally useful tokens'' may be difficult to draw sharply with behavioural experiments alone.

\paragraph{Implications for evaluation methodology.}
Single-validator replications may misestimate the verbose-helps signal by $3{-}4\times$ relative to dual-validator estimates.
We recommend that future CoT mechanism studies report both algorithmic and LLM-judge validators side by side, and disclose cross-judge agreement ($\kap$) as a routine metric.

\paragraph{Open questions and scope.}
Our design cannot quantify the lower bound of any forward-pass contribution---the evidence is that a purely length-based forward-pass account is insufficient, not that forward-pass effects are zero.
Verbose-helps is also not universal across reasoning domains: the oracle-trace experiment (\Cref{app:pilot}) found that on StrategyQA's commonsense reasoning, verbose CoT \emph{hurts} accuracy for 2 of 4 model families, suggesting that the benefits documented in this paper may be specific to mathematical and symbolic reasoning.

% =============================================================================
% 6. LIMITATIONS
% =============================================================================
\section{Limitations}
\label{sec:limitations}

\textbf{Qwen-family reliance in the intervention leg.}
In Leg~2, three of four targets, the shared rewriter, and the E3 judge are Qwen-family (only OLMo is cross-family).
The conclusions do not hinge on this: the in-distribution leg spans many non-Qwen families and finds the effect split by post-training rather than family (non-Qwen reasoners---Phi-4, Magistral, gpt-oss---are all $\Delta \approx 0$), and two non-Qwen judges leave the findings intact (Claude Opus 4.8 on the same-plan mixed pairs, \Cref{sec:samedag-results}; Yi-1.5-9B on Tier~C, \Cref{sec:tierc}).
Residual Qwen reliance is thus confined to the intervention leg's instruments, not the cross-leg conclusion.

\textbf{Local-judge agreement (intervention leg).}
For the E3 equivalence labels used in Leg~2, binary $\kap = 0.586$ between Sonnet and the local judge is moderate, not substantial.
The disagreement is structural (implicit-step completeness thresholds differ between LLMs) rather than bias-correctable, so additional calibration is unlikely to close the gap fully.
The in-distribution leg's same-plan judging is validated separately against Claude Opus 4.8 (see ``Same-plan effect size and instrument'').

\textbf{Same-plan effect size and instrument.}
The one positive same-plan effect that survives on both benchmarks---DeepSeek-R1-Distill---is small (${\sim}2$\,pp pooled) and 2 of its 5 arms lose individual significance under the LaTeX-normalised grader, so the \emph{family} claim is stronger than any single arm.
On MATH-500 the E3 judge is the entire same-plan instrument (no DAG pre-filter), which is why the independent Opus cross-check is load-bearing; GSM8K covers only ${\sim}13$ models, so the full 25-model lineage claim is MATH-500-only; and mistral-7B's underpowered MATH-500 cell is excluded from the non-reasoner pool (its well-powered GSM8K cell is retained).

\textbf{Shared-rewriter length asymmetry.}
Shared-L4 rewrites are 17--45\% longer than self-L4 at the median.
The rewriter-quality finding is therefore confounded with additional tokens from the shared source.
Within-source L4$-$L1 gap comparisons and the redaction-amplification analysis partially deconfound this, but exact token-controlled regression has not been performed.

% =============================================================================
% 7. CONCLUSION
% =============================================================================
\section{Conclusion}

Comparing a model's \emph{own} concise and verbose natural samples at a fixed reasoning plan, the apparent ``overthinking'' length penalty dissolves: across 25 models, extra tokens do essentially nothing for independently-trained reasoners, and where length does help---DeepSeek-R1-Distill's released weights, and non-reasoners on GSM8K---a blind analysis attributes the gain to validation and checking content rather than token count.
Because this leg is entirely in-distribution, it answers the central objection to supplied-trace experiments.
A controlled-intervention battery---rewriting traces to matched verbosity while holding DAG-verified content fixed, under dual validators across eight benchmarks---reaches the same conclusion: verbose-helps replicates in direction but is modest (1--4\,pp under the LLM judge versus 5--10\,pp under the algorithmic validator alone) and tracks the quality of the verbose prose rather than its length; under maximum numerical redaction it \emph{grows} $3.24\times$, and length-matched non-reasoning filler recovers none of it.
Both lines converge on a single picture: what matters is what the extra tokens \emph{do}---the reasoning and validation content they carry---not how many there are.
We recommend dual-validator reporting and cross-judge calibration as standard practice for CoT mechanism studies, alongside in-distribution same-plan comparisons that hold reasoning content fixed rather than relying on prompt-based length interventions.

% =============================================================================
% REFERENCES
% =============================================================================
\bibliography{references}
\bibliographystyle{plainnat}

% =============================================================================
% APPENDIX
% =============================================================================
\clearpage
\appendix
\begin{center}
\Large\textbf{Appendix}
\end{center}
\vspace{0.5em}

\section{Local Judge Calibration}
\label{app:calibration}

\begin{figure}[h]
\centering
\includegraphics[width=0.7\columnwidth]{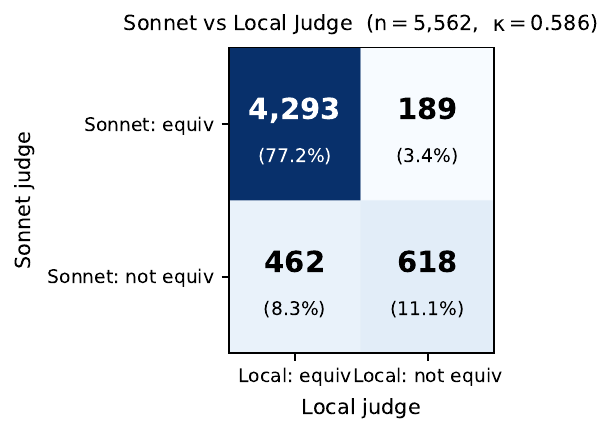}
\caption{Binary confusion matrix: Sonnet versus local Qwen judge on 5{,}562 GSM8K rows ($\kap = 0.586$, agreement = 88.3\%).}
\label{fig:calibration}
\end{figure}

\Cref{fig:calibration} shows the binary confusion matrix between Sonnet and the local Qwen3-Next-80B judge on 5{,}562 jointly labelled GSM8K rows.
The dominant agreement cell (Sonnet = equivalent, Local = equivalent) accounts for 77.2\% of rows.
The main disagreement pattern is the local judge classifying rewrites as equivalent that Sonnet classifies as not-equivalent (8.3\%), suggesting the local judge is slightly more permissive on implicit-step completeness.

\section{Full Result Tables}
\label{app:tables}

\Cref{tab:e2full,tab:e3full} report the complete $\DeltaEM$ panels with 95\% bootstrap confidence intervals and paired sample sizes.

\begin{table}[t]
\centering
\caption{$\DeltaEM$ under E2 (algorithmic validator). Format: $\Delta$ [CI$_\text{lo}$, CI$_\text{hi}$] ($n$). Tier~C undefined under E2.}
\label{tab:e2full}
\resizebox{\textwidth}{!}{%
\begin{tabular}{@{}lcccc@{}}
\toprule
& OLMo-7B & Qwen3-4B & Qwen3-30B & Qwen3-32B \\
\midrule
GSM8K & $+$.069 [$+$.047, $+$.094] (532) & $+$.053 [$+$.037, $+$.069] (688) & $+$.021 [$+$.004, $+$.037] (625) & $+$.031 [$+$.012, $+$.053] (466) \\
MATH-500 & $+$.039 [$-$.032, $+$.113] (100) & $+$.088 [$+$.034, $+$.142] (146) & $+$.078 [$+$.032, $+$.129] (153) & $+$.017 [$-$.031, $+$.065] (108) \\
MultiArith & $+$.023 [$-$.002, $+$.054] (111) & $+$.022 [$-$.003, $+$.050] (113) & $-$.014 [$-$.037, $+$.001] (95) & $-$.001 [$-$.022, $+$.019] (98) \\
SVAMP & $+$.021 [$+$.005, $+$.043] (137) & $+$.016 [$-$.005, $+$.041] (151) & $+$.003 [$-$.011, $+$.020] (146) & $+$.008 [$-$.016, $+$.038] (118) \\
AQuA & $+$.000 [$+$.000, $+$.000] (18) & $+$.006 [$-$.030, $+$.042] (21) & $+$.042 [$-$.005, $+$.106] (21) & $+$.097 [$+$.021, $+$.215] (8) \\
\midrule
Letter & \multicolumn{4}{c}{\textit{E2 undefined (non-arithmetic)}} \\
Coin & \multicolumn{4}{c}{\textit{E2 undefined (non-arithmetic)}} \\
StrategyQA & \multicolumn{4}{c}{\textit{E2 undefined (non-arithmetic)}} \\
\bottomrule
\end{tabular}%
}
\end{table}

\begin{table}[t]
\centering
\caption{$\DeltaEM$ under E3 (LLM judge). Format: $\Delta$ [CI$_\text{lo}$, CI$_\text{hi}$] ($n$).}
\label{tab:e3full}
\resizebox{\textwidth}{!}{%
\begin{tabular}{@{}lcccc@{}}
\toprule
& OLMo-7B & Qwen3-4B & Qwen3-30B & Qwen3-32B \\
\midrule
GSM8K & $+$.027 [$+$.018, $+$.037] (1257) & $+$.014 [$+$.004, $+$.024] (1246) & $-$.019 [$-$.031, $-$.008] (1221) & $+$.000 [$-$.012, $+$.013] (1197) \\
MATH-500 & $+$.069 [$+$.038, $+$.102] (268) & $+$.025 [$-$.006, $+$.052] (273) & $+$.005 [$-$.021, $+$.030] (273) & $+$.018 [$-$.009, $+$.048] (248) \\
MultiArith & $+$.014 [$+$.002, $+$.029] (143) & $+$.002 [$-$.011, $+$.014] (144) & $-$.032 [$-$.056, $-$.013] (143) & $-$.006 [$-$.028, $+$.016] (143) \\
SVAMP & $+$.027 [$+$.009, $+$.048] (185) & $+$.020 [$+$.005, $+$.037] (189) & $-$.004 [$-$.015, $+$.008] (185) & $+$.028 [$+$.007, $+$.052] (185) \\
AQuA & $-$.000 [$-$.005, $+$.003] (231) & $-$.004 [$-$.016, $+$.007] (229) & $-$.010 [$-$.026, $+$.005] (231) & $+$.007 [$-$.009, $+$.026] (224) \\
\midrule
Letter & $+$.276 [$+$.248, $+$.302] (457) & $+$.030 [$+$.016, $+$.044] (455) & $+$.047 [$+$.028, $+$.065] (452) & $+$.040 [$+$.023, $+$.056] (443) \\
Coin & $+$.026 [$+$.006, $+$.044] (489) & $-$.021 [$-$.035, $-$.007] (488) & $-$.206 [$-$.224, $-$.188] (487) & $+$.164 [$+$.145, $+$.184] (484) \\
StrategyQA & $-$.026 [$-$.032, $-$.019] (2185) & $+$.023 [$+$.018, $+$.029] (2181) & $-$.023 [$-$.029, $-$.016] (2156) & $+$.057 [$+$.049, $+$.065] (2112) \\
\bottomrule
\end{tabular}%
}
\end{table}

\section{Worked Example: One Pipeline Pass}
\label{app:example}

\Cref{fig:worked} illustrates the full pipeline for one GSM8K problem.
The example shows how the same anchor reasoning is rewritten at two verbosity levels, validated, number-redacted, and used to prompt target completion.

\begin{figure}[t]
\small
\begin{tabular}{@{}p{0.47\textwidth}p{0.47\textwidth}@{}}
\toprule
\multicolumn{2}{@{}l}{\textbf{Question:} He had 6 toys, then bought 8 more, gave away 5, then doubled the remainder. How many toys does he have?} \\
\multicolumn{2}{@{}l}{\textbf{Gold answer:} 18} \\
\midrule
\multicolumn{2}{@{}l}{\textbf{Anchor trace} (target model's original reasoning):} \\
\multicolumn{2}{@{}p{0.95\textwidth}}{He had 6 toys, then bought 8 more (6 + 8 = 14), then gave away 5 (14 $-$ 5 = 9), then doubled to (9 $\times$ 2 = 18). So the answer is 18.} \\
\midrule
\textbf{L1 (concise) rewrite:} & \textbf{L4 (verbose) rewrite:} \\
\begin{minipage}[t]{0.45\textwidth}\vspace{0pt}
\texttt{6 + 8 = 14}\\
\texttt{14 - 5 = 9}\\
\texttt{9 * 2 = <REDACTED>}
\end{minipage}
&
\begin{minipage}[t]{0.45\textwidth}\vspace{0pt}
Starting from the initial count, we have six toys. He then acquires eight more, which by simple addition gives him a total of six plus eight equals fourteen toys. Subsequently, five toys are given away, leaving fourteen minus five equals nine toys. Finally, the count is doubled, yielding nine times two equals \texttt{<REDACTED>}.
\end{minipage} \\
\midrule
\multicolumn{2}{@{}l}{\textbf{Number-redacted completion prompt} (shown for L4):} \\
\multicolumn{2}{@{}p{0.95\textwidth}}{\texttt{Q: He had 6 toys, then bought 8 more, gave away 5, then doubled the remainder. How many toys does he have?}} \\
\multicolumn{2}{@{}p{0.95\textwidth}}{\texttt{A: Starting from the initial count, we have <REDACTED> toys. He then acquires <REDACTED> more, which by simple addition gives him a total of <REDACTED> plus <REDACTED> equals <REDACTED> toys. Subsequently, <REDACTED> toys are given away, leaving <REDACTED> minus <REDACTED> equals <REDACTED> toys. Finally, the count is doubled, yielding <REDACTED> times <REDACTED> equals <REDACTED>.}} \\
\multicolumn{2}{@{}l}{\texttt{The answer is}} \\
\bottomrule
\end{tabular}
\caption{Worked example of one pipeline pass. The anchor is rewritten at two verbosity levels (L1 concise, L4 verbose). The DAG---three operations in dependency order---is identical in both rewrites. Numeric tokens are replaced with \texttt{<REDACTED>} before the target model completes ``The answer is \underline{\hspace{1cm}}''. The question asks whether the target performs better after reading the verbose (L4) trace than the concise (L1) trace.}
\label{fig:worked}
\end{figure}

\paragraph{E3 judge prompt (excerpt).}
The LLM judge receives a per-benchmark rubric as the system message, followed by the (anchor, rewrite) pair.
For GSM8K, the rubric specifies: ``Same set of arithmetic operations (same numbers combined with same operators to produce the same intermediate values). Same dependency structure. Same final answer.''
A \texttt{REDACTED}-aware clarification block is appended: ``A rewrite with a redacted final answer IS equivalent if the reasoning operations and dependency structure that PRODUCE that answer are preserved.''
The judge output is constrained to a JSON object: \texttt{\{``verdict'': ``equivalent'' | ``not\_equivalent'' | ``ambiguous'', ``confidence'': [0,1], ``rationale'': ``...''\}}.

\section{Additional Experiments: Oracle-Trace Verbosity Ladder}
\label{app:pilot}

The main experiment constructs traces from the target model's own reasoning, then validates and redacts them before completion.
Here we take a complementary approach: since directly sampling from the model's natural CoT distribution while controlling verbosity is difficult, we instead start from an \emph{oracle trace}---a canonical reasoning DAG extracted from the gold solution for each test problem---and ask Qwen3-32B to rewrite it into four verbosity levels (L1--L4, mean 19--285 words on GSM8K).
Each rewritten trace is then supplied as a prefix to the target model: the model receives the question followed by the reasoning trace (truncated before the final derivation to prevent answer copying) and completes ``The answer is \underline{\hspace{0.5cm}}''.
This prefix-completion setting directly measures $p(\text{answer} \mid \text{question}, \text{trace})$ at varying verbosity while holding the underlying reasoning graph fixed.
We also include a \emph{filler} condition---the L1 trace padded with fluent but non-reasoning natural-language text to match the L4 token count---to test whether raw token count alone accounts for any verbose-helps effect.
We evaluate 6 targets across 4 model families on GSM8K ($n \approx 1100$ per condition) and 4 targets on StrategyQA ($n = 2287$).

This design is simpler and more direct than the main experiment's pipeline: it uses oracle traces rather than model-generated anchors, omits number-redaction and dual-validator filtering, and evaluates prefix-completion accuracy with a single answer-match check.
In this respect it is closer to the experimental setting of \citet{jin2024impact}, who vary verbosity in few-shot CoT demonstrations and measure downstream accuracy---though we supply the reasoning trace for the target problem directly rather than modifying few-shot examples.

\begin{figure}[h]
\centering
\includegraphics[width=\columnwidth]{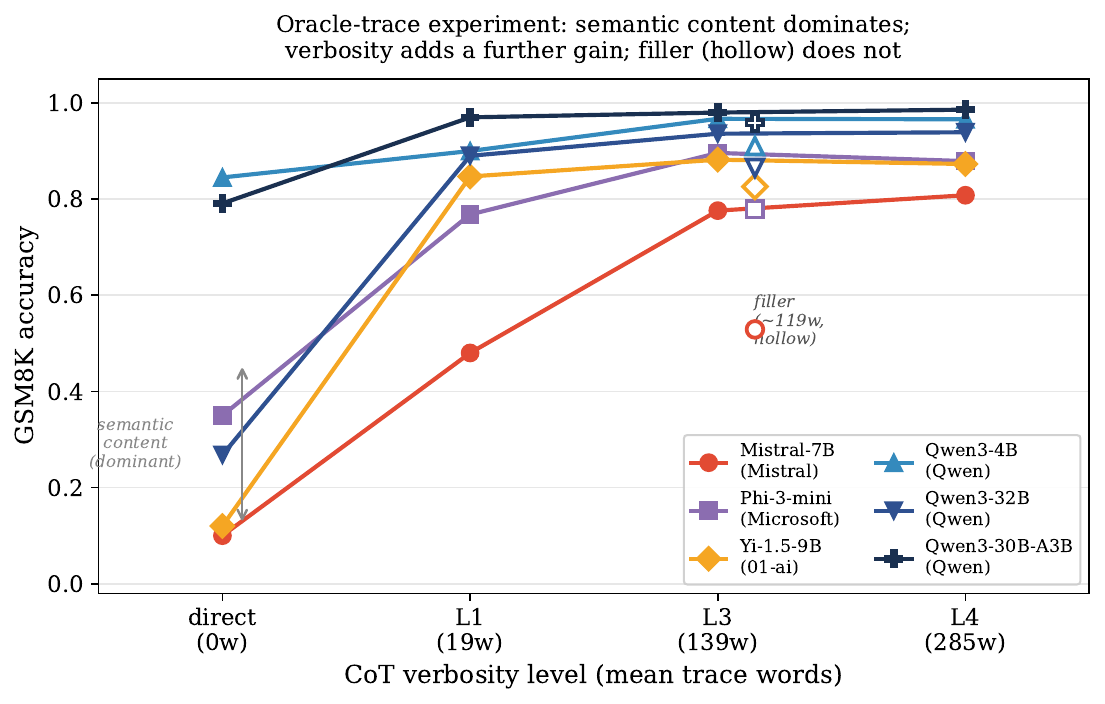}
\caption{GSM8K accuracy at direct-answer, L1 (${\sim}19$\,w), L3 (${\sim}139$\,w), and L4 (${\sim}285$\,w) for 6 models across 4 families. Hollow markers show the filler condition (${\sim}119$\,w, non-semantic padding), plotted at the L3 position since filler and L3 are closest in length. Filler clusters below L3, showing that extra tokens without semantic content do not recover the verbose-helps effect.}
\label{fig:pilot_dose}
\end{figure}

\paragraph{Dose-response across four families.}
\Cref{fig:pilot_dose} shows GSM8K accuracy at direct-answer, L1, L3, and L4 for 6 targets from 4 families, with filler (${\sim}119$\,w) shown as hollow markers at the L3 position.
The dominant feature is the massive jump from direct-answer to L1 ($+5$ to $+73$\,pp), confirming that the semantic content of CoT does most of the work.
Going from L1 to L4 adds a smaller gain, with magnitudes scaling inversely with the model's L1 baseline (Mistral-7B: $+33$\,pp from a 48\% baseline; Qwen3-30B-A3B: $+2$\,pp from a 97\% baseline).

\paragraph{Length-matched filler does not recover the CoT benefit.}
This experiment included a token-matched filler condition: the concise L1 trace padded with fluent but non-reasoning natural-language text to match the L4 token count.
\Cref{tab:pilot_filler} shows that filler$-$L1 is near zero on all six models, while L3$-$filler---comparing two conditions of similar length (${\sim}119$\,w vs ${\sim}139$\,w) where only the semantic content differs---is significantly positive on every model ($+2$ to $+25$\,pp).
Adding tokens without semantic reasoning content does not reproduce the verbose-helps effect.

\begin{table}[h]
\centering
\small
\caption{Filler analysis (oracle-trace experiment, GSM8K). Filler$-$L1 tests whether extra tokens alone help; L3$-$filler tests whether semantic content at matched length helps. $*$ = 95\% bootstrap CI excludes zero.}
\label{tab:pilot_filler}
\begin{tabular}{@{}lcc@{}}
\toprule
Model & filler$-$L1 & L3$-$filler \\
\midrule
Mistral-7B & $+$0.049$*$ & $+$0.247$*$ \\
Phi-3-mini & $+$0.011 & $+$0.117$*$ \\
Yi-1.5-9B & $-$0.021 & $+$0.056$*$ \\
Qwen3-4B & $+$0.010 & $+$0.057$*$ \\
Qwen3-32B & $-$0.025$*$ & $+$0.071$*$ \\
Qwen3-30B-A3B & $-$0.011$*$ & $+$0.021$*$ \\
\bottomrule
\end{tabular}
\end{table}

\begin{figure}[h]
\centering
\includegraphics[width=\columnwidth]{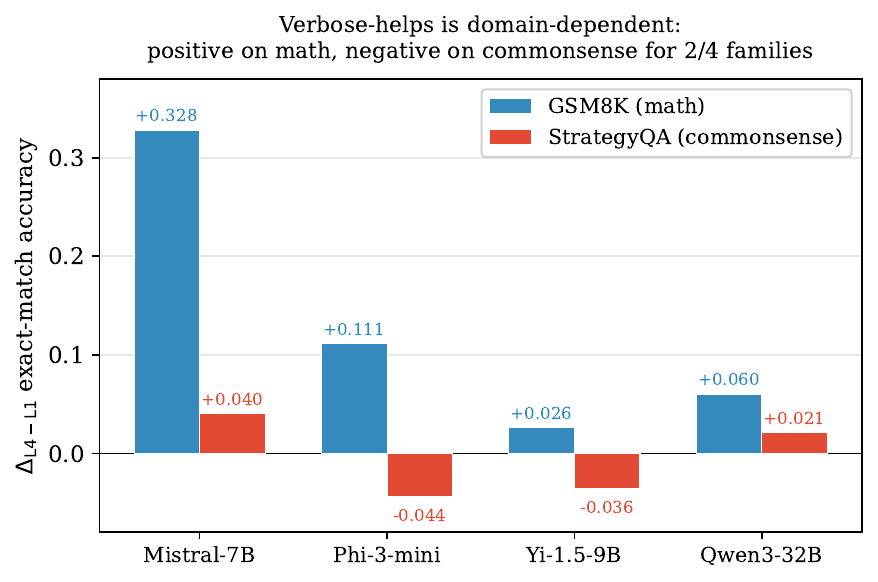}
\caption{$\DeltaEM$ (L4$-$L1) on GSM8K (math, blue) versus StrategyQA (commonsense, red) for 4 models. Verbose-helps is positive on math for all models, but reverses on commonsense for Phi-3-mini and Yi-1.5-9B.}
\label{fig:pilot_domain}
\end{figure}

\paragraph{Domain-dependent verbose-helps.}
\Cref{fig:pilot_domain} compares $\DeltaEM$ (L4$-$L1) on GSM8K versus StrategyQA for 4 models.
On GSM8K (math reasoning), all four models show positive verbose-helps.
On StrategyQA (commonsense reasoning), 2 of 4 families show the \emph{opposite} pattern: Phi-3-mini ($\Delta_{\text{L4--L1}} = -0.044$) and Yi-1.5-9B ($-0.036$) are significantly \emph{hurt} by verbose CoT.
This domain-dependence constrains the generality of verbose-helps claims.

\paragraph{Corruption universally hurts.}
Swapping a correct reasoning step for an incorrect one degraded EM by 5--20\,pp across every tested target on GSM8K ($\approx$5--7\,pp on the Qwen targets, up to $\approx$20\,pp on Mistral, OLMo, and Phi-3).
On StrategyQA, corrupted CoT performed \emph{catastrophically} worse than no CoT at all ($-9$ to $-34$\,pp below direct-answer), with corrupted accuracy near chance (${\sim}47$--$50\%$ on a yes/no task).
This replicates under the main experiment's stricter methodology, confirming that the target model conditions on the semantic content of the supplied trace.

\paragraph{Compression to ${\sim}27$ words preserves accuracy.}
Self-trace compression---reducing a target model's own 300-word reasoning to approximately 27 words while preserving the semantic DAG---produced no detectable EM cost ($n = 300$, all pairwise contrasts statistically null).
Combined with the main experiment's L1--L4 comparison, this maps out a broad accuracy plateau from ${\sim}27$ words through L4.

\paragraph{Prefill equivalence.}
This experiment established that prefilling a fixed trace and teacher-forcing that same trace token-by-token produce essentially identical target conditioning ($|\Delta| \leq 0.17$ nats per token, $n = 50$ problems $\times$ 5 conditions).
This foundational result underwrites the main experiment's measurement step, which prefills the question and rewrite before reading the target's greedy continuation.

\end{document}